\documentclass[final]{cvpr}

\usepackage{times}
\usepackage{epsfig}
\usepackage{graphicx}
\usepackage{amsmath}
\usepackage{amssymb}
\usepackage{multirow}

\usepackage[pagebackref=true,breaklinks=true,colorlinks,bookmarks=false]{hyperref}

\def\cvprPaperID{****} 
\def\confYear{CVPR 2021}

\begin{document}

\title{Self-Supervised Learning with Fully Convolutional Networks}

\author{Zhengeng Yang~~~Hongshan Yu\thanks{Corresponding author}~~~Yong He\\
Hunan University\\
{\tt\small \{yzg050215,Yuhongshan,h.yong\}@hnu.edu.cn}
\and
Zhi-Hong Mao\\
University of Pittsburgh\\
{\tt\small zhm4@pitt.edu}
\and
Ajmal Mian\\
University  of Western Australia\\
{\tt\small ajmal.mian@uwa.edu.au}
}

\maketitle

\begin{abstract}
  Although deep learning based methods have achieved great success in many computer vision tasks, their performance relies on a large number of densely annotated samples that are typically difficult to obtain. In this paper, we focus on the problem of learning representation from unlabeled data for semantic segmentation. Inspired by two patch-based methods, we develop a novel self-supervised learning framework by formulating the Jigsaw Puzzle problem as a patch-wise classification process and solving it with a fully convolutional network. By learning to solve a Jigsaw Puzzle problem with 25 patches and transferring the learned features to semantic segmentation task on Cityscapes dataset, we achieve a 5.8 percentage point improvement over the baseline model that initialized from random values. Moreover, experiments show that  our  self-supervised  learning  method can be applied to different datasets and models.  In paticular, we achieved competitive performance with the state-of-the-art methods on the PASCAL VOC2012 dataset using significant fewer training images.
\end{abstract}


\section{Introduction}

In recent years, deep convolutional neural networks (CNN) have been advancing frontiers of many computer vision tasks such as image classification~\cite{he2016deep,sandler2018mobilenetv2}, semantic segmentation~\cite{yang2020ndnet,yang2020small, yu2018methods,zhao2017pyramid}. However, the performance of deep CNNs relies heavily on large amounts of labeled data. Data labeling requires intensive manual effort and is not even feasible for some applications. As a result,  learning deep representation from unlabeled data has recently received great attention~\cite{feng2019self}. A promising strategy in this line of research is self-supervised learning which utilizes automatically generated labels for supervision. For example, Gidaris et al.~\cite{gidaris2018unsupervised} learn image representation by training an image rotation aware network. Note that many tasks (e.g., rotation prediction) in self-supervised learning is usually not the target task of interest, however, the representations learned during the process are still very useful. Such tasks are generally referred to as proxy or surrogate tasks in the current literature.

Many proxy tasks for representation learning have been proposed over the last few years. One such popular task is to exploit the spatial context within the visual images as the supervisory signal. For example, Doersch et al.~\cite{doersch2015unsupervised} cropped a pair of neighbor patches and trained a network to predict their relative locations from the eight possible options. However, since the sampled patches had small sizes (e.g., $96\times96$), it is easy for one of the two patches to cover an area that contains little or no useful information, especially in high resolution images. Using more patches could be better choice for this problem, 
for example, the jigsaw puzzle system~\cite{noroozi2016unsupervised} 
used nine $80\times80$ patches. However, to extract features from patches, these methods adopt the Siamese network architecture where the training time increases significantly with the increase in the number of patches used.

Another drawback of current patch-based methods is that most of them use image classification as the main target task. Hence, they usually perform self-supervised learning on large-scale image classification datasets, such as the ImageNet~\cite{deng2009imagenet}, and then transfer the learned weights to other tasks e.g., object detection.
However, such an approach is sub-optimal due to two reasons. Firstly, training on a large-scale image classification dataset is time consuming, especially when graphic processing unit (GPU) memories are limited. For example, it takes four weeks for Doersch et al.~\cite{doersch2015unsupervised} to train their models on ImageNet. 
Secondly, features learned for image classification may not be suited for other tasks due to the difference in the data distribution.
Given that we can easily access massive amounts of unlabeled data for most applications, we believe that it is better to perform self-supervised learning on the same scenes of the target task to avoid to domain gap problem.

To achieve the goal of self-supervised learning of feature representations for semantic segmentation, we incorporate the idea of Jigsaw ~\cite{noroozi2016unsupervised} 
and relative location prediction ~\cite{doersch2015unsupervised} into a unified framework in this paper. More specifically, we select nine patches to fully exploit the spatial contexts for representation learning similar to Jigsaw. However, instead of predicting the permutation of nine patches, we propose to predict the relative locations between the central patch and its eight neighbors at a time. This reduces the number of parameters in the last layer of Jigsaw by 4.5 times. Highlights and contributions of this paper are as follows.
\begin{enumerate}
\item We demonstrate 
that fully convolutional networks (FCN)~\cite{shelhamer2017fully} can be approximately viewed as patch-wise classification networks.
\item Based on the above, we propose a self-supervised learning framework that solves the Jigsaw Puzzle problem with a FCN. By transferring our self-supervised models to the semantic segmentation task, we achieved a 2.8\% point improvement on the mIoU compared with the baseline initialized with random values.
\item 
Using our proposed FCN-based method which requires fewer parameters, we increase the number of Jigsaw patches in self-supervised learning to 25 and further improve the representation learning power.
\end{enumerate}


The rest of this paper is organized as follows. The next section presents related work in self-supervised learning methods.  Section \ref{sec_method} describes the theoretical analysis and implementation of our method in detail. Comprehensive ablation studies and experimental analysis are conducted in Section \ref{sec_exp}.  Finally, conclusions are presented in Section \ref{sec_conclusion}.



\begin{figure*}[ht]
\centering\includegraphics[width=6.5in]{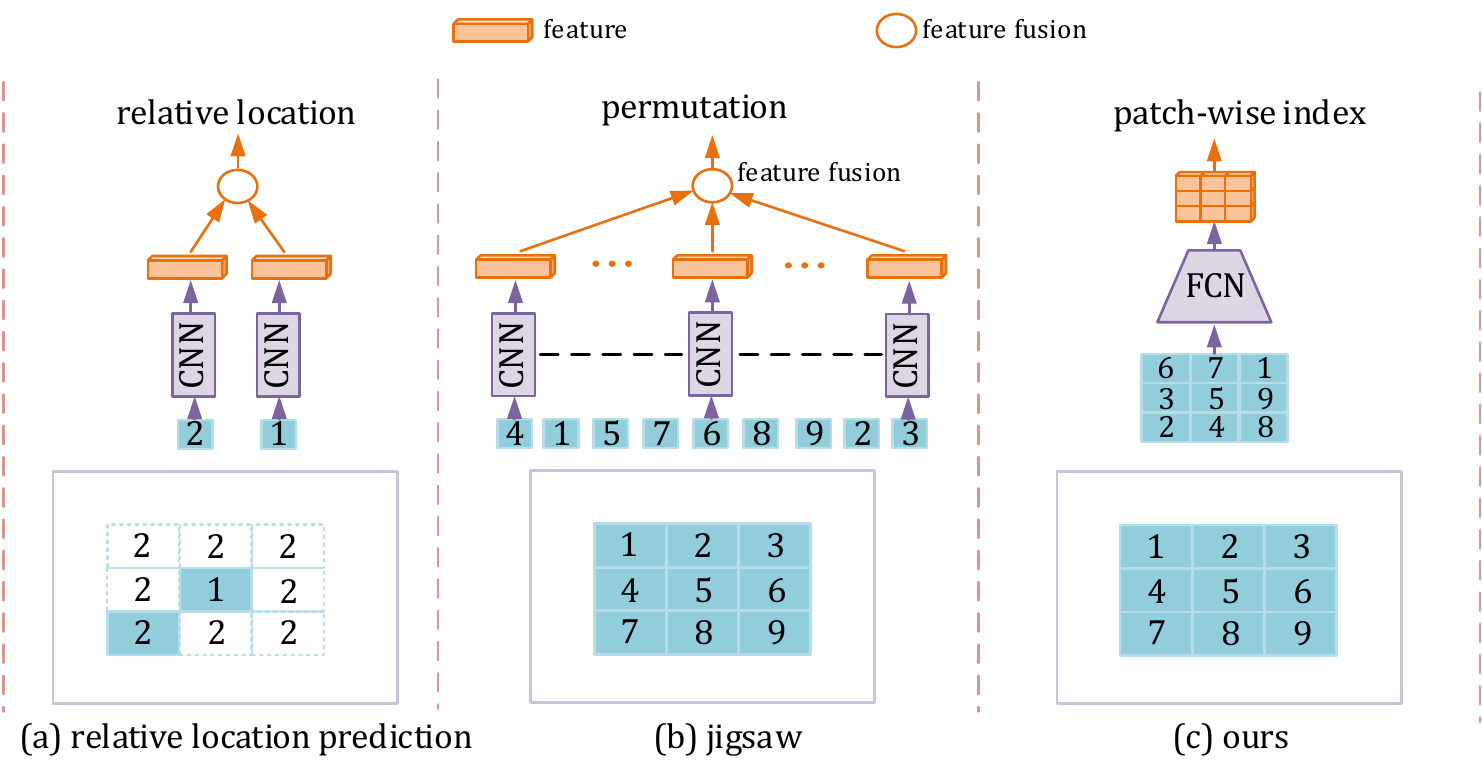}
\caption{
Current patch-based methods adopt the Siamese network to extract features from each patch independently. This leads to a bloated classification layer. Our method adopts a FCN to extract features from all patches simultaneously. This allows us to extend the number of patches to 25 without increasing the network complexity for better transfer learning.}
\label{fig_oursvsothers}
\end{figure*}


\section{Related Work}
\label{sec_related}
We organize our survey of the existing literature based on the proxy tasks adopted for self-supervised learning. Based on the proxy tasks, current self-supervised learning methods, that are most relevant to our work, can be roughly divided into two large categories namely, context prediction and image generation.

Methods in the context prediction category try to exploit the internal spatial context within the visual data for supervisory signal. For example, Doersch et al. ~\cite{doersch2015unsupervised} used the eight possible relative locations between a pair of neighbor patches as the label for patch classification. Noroozi and Favaro~\cite{noroozi2016unsupervised} extended this idea of relative location prediction to solve a Jigsaw Puzzle problem with nine patches. The nine patches are shuffled randomly and then the Jigsaw is performed with a network to recover the orders of them. Since the possible permutations of nine numbers is up to $9!=362,880$, which maybe infeasible to discriminate with deep network. Thus, a set of predefined permutations (supposing 100) is used for the random patch shuffling. Then, the Jigsaw is transformed into a 100-classes classification problem. Later, Mundhenk et al.~\cite{nathan2018improvements} improved these patch-based methods by incorporating numerous tricks, such as harnessing 
the jitters applied to the patches.

The image generation based methods can be divided into two stages. In the first stage, part of an image is removed. The second stage then tries to recover or generate the removed part. For example, Pathak et al.  ~\cite{pathak2016context} manually removed a region from an image and then trained a network to perform inpainting to recover the removed region using information from the remaining pixels.
Another example of image generation is the commonly used automatic colorization task, i.e., recovering the three color channels from the remaining channel(s). Zhang et al.~\cite{zhang2016colorful} and Larsson et al.~\cite{larsson2016learning} were among the first to use image colorization as a proxy task for representation learning. In particular,  Zhang et al.~\cite{zhang2016colorful} generate color channels from luminance, which is achieved by quantifying the Lab color space into a number of discrete intervals and then formulating the image colorization as a classification problem. Later the same authors~\cite{zhang2017split} extended their idea to perform cross-channel generation, i.e., the remove/generate occurs between each pair of channels. Concurrently, Larsson et al.\cite{larsson2017colorization} replaced the commonly used AlexNet~\cite{krizhevsky2012imagenet} with VGG~\cite{simonyan2014very} and introduced the hypercolumns~\cite{hariharan2015hypercolumns} to improve the learning capacity of the colorization network. Compared with patch-based methods, colorization-based methods have an advantage that they do not change the spatial structure of the input. However, colorization-based method may not not learn color based features which is important for many tasks such as semantic segmentation.

Besides the two categories mentioned above, there are many other proxy tasks designed for self-supervised learning. For example, Dosovitskiy et al.~\cite{dosovitskiy2015discriminative} defined a series of exemplar classes for representation learning, where samples of each class were generated by applying different transformations to an image patch containing an obvious object. Gidaris et al.~\cite{gidaris2018unsupervised} defined four angler options (0,90,180,270) for image rotation prediction. However, for the semantic segmentation dataset, it is difficult to meet this prerequisite since there are usual many objects of different classes in an image.
More recently, Jenni and Favaro~\cite{jenni2018self} suggested that discriminative features can be learned by distinguishing real images from images
with synthetic artifacts.

Whereas the above mentioned methods are mostly 
designed for static images, there are also many methods that leverage videos for self-supervised learning.
Typical proxy tasks the exploit video data include temporal order prediction~\cite{lee2017unsupervised,misra2016shuffle}, and future frame prediction~\cite{srivastava2015unsupervised}.



Among these works, the work most related to our method are the relative location prediction~\cite{doersch2015unsupervised} and Jigsaw method~\cite{noroozi2016unsupervised}.  As mentioned previously, the relative location prediction considered spatial contexts between only two patches of small sizes at one time, which may lead to one of them contains little useful information. In contrast, our method exploit contexts from nine or even 25 patches. The Jigsaw method also considered nine patches. However, it extracts features for each patch using the same network and then concatenates features of these nine patches for permutation classification, which leads to the final layer highly bloated. For example, if the feature channels of each patch is set to $512$, the final feature layer will contain $512\times9=4608$ channels. Then, it will requires $1\times1\times4608\times512\approx 2.36M$ parameters if we want to reduce the feature dimension to 512 for final classification. In contrast, we set the central patch as reference patch and then concatenate its feature to other patches for relative location prediction, which requires about $0.52M$ parameters using the above example. Moreover, our FCN-based network is also different from the network used in Jigsaw.

\section{Method}
\label{sec_method}
\subsection{Fully Convolutional Network}

Given an image, typical CNNs convert it to a single feature vector and then predict its class. This is achieved by multiple ``convolution+pooling" blocks and a separate spatial dimension reduction 
operation that usually presents as a fully connected layer (e.g., VGG~\cite{simonyan2014very}) or global average layer (e.g., DenseNet~\cite{huang2017densely}). To produce dense predictions, the FCN~\cite{shelhamer2017fully} discards the spatial dimension reduction operation and then make prediction at each spatial location by applying a softmax function as follows
\begin{equation}
p_{c|X_{(i,j)}}=\frac{e^{W_cX_{ij}}}{\sum_{b \in classes} e^{W_bX_{ij}}},
\label{eq_fcn}
\end{equation}
where $X_{i,j}$ is the feature vector at spatial location $(i,j)$, and $W_c$ is the classification parameters of class $c$. The production of $W_c$ and $X_{i,j}$ is also called class score for class $c$ at spatial location $(i,j)$.

\begin{figure}[ht]
\centering\includegraphics[width=3.1in]{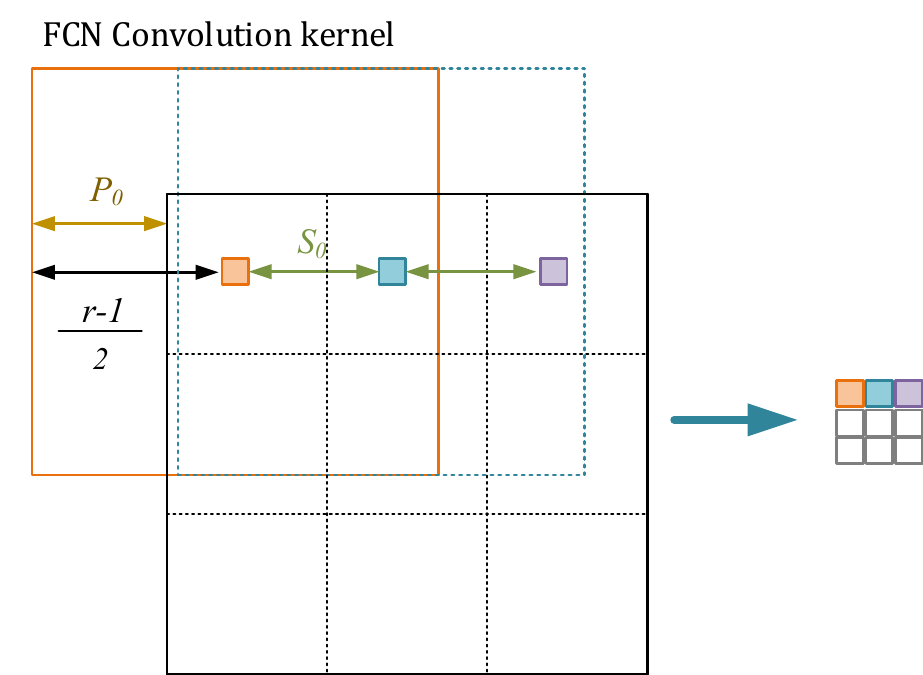}
\caption{Receptive field of FCN.}
\label{fig_fcnconv}
\end{figure}

\subsection{Patch-wise Classification with FCN}
It can be easily shown that Equation (\ref{eq_fcn}) is essentially a ``pixel"-wise classification problem if each spatial unit of the output feature map of FCN is treated as a special ``pixel". For convenience, we call these special ``pixels'' feature pixels. Next, we show that the FCN can be approximately viewed as a patch-wise classification framework by proving that the overlapping areas between input regions (or patches), where these feature pixels generated from, can be ignored to some degree. 

Consider a FCN with $L$ layers and denote the feature map of layer $l$ as $f_l$, feature pixel at spatial coordinate $(i,j)$ of $f_l$ can be denoted by $f_l(i,j)$. Theoretically, the size of the input region where the feature pixel $f_l(i,j)$ generated from is equal to the receptive field (RF) size of $f_l$. In general, the RF size of a FCN can be computed by~\cite{araujo2019computing}

\begin{equation}
r=\sum_{l=1}^L \Bigg( (k_l-1)\prod_{i=1}^{l-1}s_i \Bigg)+1
\label{eq_receptivefield}
\end{equation}
where $k_l$ is the kernel size of layer $l$, $s_i$ is the output stride with respect to its previous layer.
To obtain the exact input region corresponding to $f_l(i,j)$, we need to further compute the coordinate of the RF center. To this end, we define two auxiliary variables following~\cite{araujo2019computing}.
The first one is effective stride: $S_0=\prod_{l=1}^L s_l$, the stride of output feature map $f_L$ with respect to the input image.
The second one is effective padding: $P_0=\sum_{l=1}^L p_l \prod_{i=1}^{l-1} s_i$, the padding added to input image from the perspective of $f_L$, where $p_l$ is the padding added to $l$.
With these two auxiliary variables, the coordinate of RF center of $f_L(i,j)$ can be computed by
\begin{equation}
(h_i,v_j)=(-P_0+ \frac{r-1}{2}+i\cdot  S_0 , -P_0 + \frac{r-1}{2}+j\cdot S_0)
\label{eq_centerofreceptivefield}
\end{equation}
In other words, we can view the entire FCN as a special convolution layer whose kernel size, stride and padding equal to $r$,  $S_0$, and $P_0$, respectively (see Fig. \ref{fig_fcnconv}). While most basic CNNs (e.g., AlexNet, VGG16, ResNet) have five layers with stride equal to 2, $S_0$ of FCN build from these networks is typically equal to 32. On the other hand, according to (\ref{eq_receptivefield}), RFs of FCNs build from these commonly used networks are usually significantly larger than $2S_0$ (see Table \ref{table_rf-of-com-net}), which means that RFs of neighbor positions of FCN's output are highly overlapped. Nevertheless,  Luo et al.~\cite{luo2016understanding} have shown that the distribution of impact within the RF is asymptotically Gaussian and the effective receptive field only takes up a small fraction of the full theoretical RF. Thus, it is reasonable to
hypothesize that the effective region where each feature pixel generated from is a small patch around the center of the RF. Then, classification based on the output features of FCN can be approximately viewed as a patch-wise classification process.




\begin{table}[ht]
\renewcommand\arraystretch{1.2}

\centering
\hspace*{-0.2cm}
\begin{tabular}{cccc}
\hline
 & AlexNet &VGG16 & ResNet101\\
 \hline
RF & 195 &212 & 1027\\
$S_0$ & 32 &32 & 32\\

\hline
\end{tabular}

\vspace{1mm}
\caption{RF and effective stride of commonly used FCNs.}
\label{table_rf-of-com-net}
\end{table}


\begin{figure*}[ht]
\centering\includegraphics[width=6.7in]{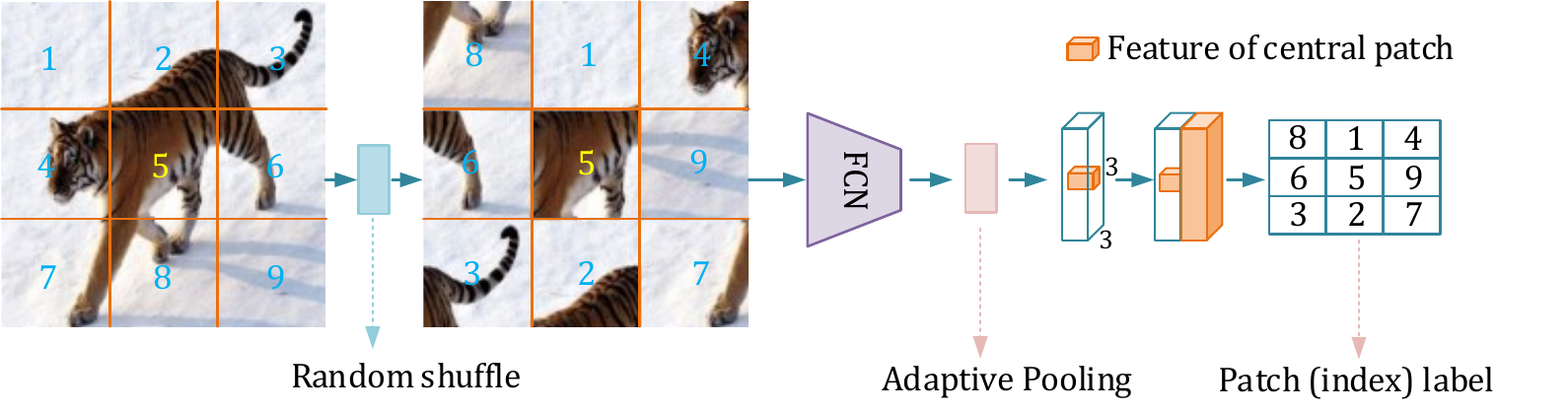}
\caption{Overview of our self-supervised learning framework. The basic idea of our self-supervised learning is to learn to predict the absolute location of patches in a Jigsaw Puzzle problem. Towards this goal, we first divide the training image into nine patches, which are then rearranged to a new image after a random shuffle operation. Then, a FCN, which we treat approximately as a patch-wise classification framework, is used to extract features for each patch. Finally, we concatenated features of each patch with features of the central patch, which act as the reference patch, for absolute location prediction.}
\label{fig_jigsawfcn}
\end{figure*}

\subsection{Self-supervised Learning with FCN}
Unlike predicting the permutations of patches (as in~\cite{noroozi2016unsupervised}), in real life, the Jigsaw Puzzle problem is typically solved by finding the absolute location of each patch. In particular, when confronted with a Jigsaw Puzzle game, a human player tries to classify each patch into the correct location according to visual information extracted from all the patches. Thus, the Jigsaw Puzzle can be viewed as a patch-wise classification problem and this is exactly the approach we take in this paper. More formally, denote the number of patches of a Jigsaw Puzzle problem as $N$, and the feature vectors of a patch $I_n$ as $f(I_n)$,  the Jigsaw Puzzle problem depicted in~\cite{noroozi2016unsupervised} can be written by
\begin{equation}
C_{I_{1'},I_{2'},...,I_{N'}}={\rm MLP}\bigg ( \big[ f(I_{1'}),f(I_{2'}),...,f(I_{N'}) \big] \bigg)
\label{eq_jigsaw1}
\end{equation}
where MLP represents a function defined by a multilayer perceptron, $[...]$ refers to the concatenation of feature vectors of different patches, $I_{n'}$ represents the $n$-th patch after a shuffle operation is performed over the patches, and $C$ is one of elements of a pre-defined permutation set. It can be concluded that Equation \ref{eq_jigsaw1} is essentially a sequence classification problem. While the Jigsaw Puzzle problem in practice is solved as
\begin{equation}
C_{I_{n'}} = {\rm MLP}\bigg( f(I_{n'}),f(I_{\overline{n}'}) \bigg)
\label{eq_jigsawpractice}
\end{equation}
where $I_{\overline{n}'}$ represents a number of patches apart from the $n'$-th patch. In this case, the label set for classification consists of N absolute positions of these patches.

Based on the above analysis on FCN and Jigsaw Puzzle problem, we propose to learn representations by solving a Jigsaw Puzzle problem with FCN. The overview of our method is shown in Fig.~\ref{fig_jigsawfcn}. First, an image is divided into a $3\times3$ grid which results in 9 non-overlapping patches. These 9 patches are then rearranged according to an order generated from a random shuffle operation. Finally, the rearranged image is fed into a FCN to predict the original position for each patch. To reduce the difficulty of the Jigsaw Puzzle problem, inspired by the relative location prediction in~\cite{doersch2015unsupervised}, we fix the central patch and use it as the reference for absolute location prediction of other patches. Thus, our FCN-based Jigsaw Puzzle can be formally written as
\begin{equation}
C_{I_{n'}} = {\rm MLP}\bigg( f(I_{n'}),f(I_{\lceil \frac{N}{2} \rceil}) \bigg)
\label{eq_jigsawours}
\end{equation}
Hence, our method can be also viewed as a relative location prediction problem. However, different from~\cite{doersch2015unsupervised}, which outputs only one relative location, our method outputs eight relative locations at one time.

\subsection{Network and Loss}
Many state of the art CNN architectures exist (e.g., ResNet~\cite{he2016deep}, DenseNet~\cite{huang2017densely}) for representation learning. However, most CNNs require large batch size for training to ensure good performance of the batch normalization~\cite{ioffe2015batch}. This leads to a huge consumption on GPU memories (especially for complex models like ResNet). Limited by the GPU resources at hand,  we adopt a well-known light-weight model named MobileNetV2~\cite{sandler2018mobilenetv2} as the backbone for feature extraction.  MobileNetV2  has  many  versions of different widths controlled by a parameter called ``width multiplier”. To use a large batch size for training, we set the width multiplier to 0.75 and further reduce the width of the last layer from 1,280 to 512.

Similar to most FCN-based methods, we define the training loss using the cross entropy between prediction and ground truth. More formally, the training loss of our 9 patch Jigsaw Puzzle problem is defined as

\begin{equation}
L=\frac{1}{9} \sum_{i=1}^{9}\underbrace{- log \frac{e^{W_{c}X_i}}{\sum_{b=1}^{9}e^{W_{b}X_i}}}_{cross\_entropy}.
\label{eq_trainingloss}
\end{equation}
where $c$ is the truth location of patch $i$ and  $X_i$ denotes the final feature of patch $i$.

\subsection{Datasets}
For practical problems, we generally have only a small number of fully annotated samples for network training. On the other hand, massive unlabeled images are easily accessible from the Internet. Thus, we propose to use self-supervised learning to learn feature representations from the unlabeled data and then apply the learned representations for the downstream task in the same scene.

\textbf{Cityscapes} Cityscapes~\cite{cordts2016cityscapes} is an urban scenes dataset sampled from 50 European cities. It has 2,975, 500, 1,525 fully annotated images for training, validation and testing, respectively. To imitate the real cases where fully annotated images are usually limited to a small number, we choose only 503 images (taken from Tubingen, Ulm, Weimar and Zurich) from the 2,975 training images to train the segmentation network, and use the other 2,476 images 
for self-supervised learning.

\textbf{}

\vspace{-3mm}
\section{Experiments}
\label{sec_exp}

\begin{table*}[ht]
\small%

\centering
\hspace*{-0.2cm}
\begin{tabular}{cc|cc|cc|cc|cc|cc}
\hline

\multicolumn{2}{c}{Self-supervised learning} &\multicolumn{10}{c}{Semantic Segmentation}\\
\hline
\multirow{2}{*}{Steps} & \multirow{2}{*}{ACC} &\multicolumn{2}{c}{block12345}  &\multicolumn{2}{c}{block2345} &\multicolumn{2}{c}{block345}&\multicolumn{2}{c}{block45}&\multicolumn{2}{c}{block5}\\
        & &random & finetune   &random & finetune &random & finetune &random & finetune &random & finetune\\

\hline
20K &60.6  &42.0  &43.4   &42.1  &\textbf{44.7}    & 41.9 & 44.0          & 41.0 &41.9 & 26.8 &27.5 \\
30K &85.1  &42.0  &43.2   & 42.0 & 44.3  &42.2  &\textbf{44.8}   & 42.2 &43.2 &26.8  &27.2 \\
50K & 92.8 &42.0 &42.8   &42.6  & \textbf{44.6}  & 42.1 &43.4       & 42.4 &43.0 &26.5  & 27.0\\
\hline
\end{tabular}
\vspace{1mm}
\caption{Ablation study of our self-supervised learning for semantic segmentation on the Cityscapes dataset. The blocks represent the 5 convolution layers of CNN and the values are mIoU. For example, ``block45'' means block4 and block5 parameters are learned after initializing from random values and block1 to block3 parameters are transferred from the proxy task of solving the Jigsaw puzzle problem.}
\label{table_ablationstudycity}
\vspace{-3mm}
\end{table*}
\par

\subsection{Self-supervised Learning as Target Task}
\label{subsec_selfastarget}
We first treat our self-supervised learning as the target task to prove that the Jigsaw Puzzle problem can be solved by a FCN.
Each sample of the training batch is a $576\times576$ image patch randomly sampled from an image augmented from original Cityscapes dataset. The augmentations include random mirroring and random scaling that are commonly used in deep learning. We set the batch size to 36 and the training iterations to 50K. The learning rate is set to 0.1 and divided by 5 at 10K iteration and then decayed by a factor of 10 every 10K iterations. 
We use the stochastic gradient descent (SGD) with momentum of 0.9 for parameter optimization.

Since our self-supervised learning is designed to solve a patch-wise classification task, we use the classification accuracy to evaluate its performance. Evaluation is performed over the 500 validation images of Cityscapes dataset. Note that we sampled only one patch for each image during this process, which means that we performed the Jigsaw task only 500 times during testing.

Training and evaluation were both implemented in Pytorch~\cite{paszke2019pytorch} and conducted on a computer equipped with one Titan X GPU. The experimental results are shown in Table \ref{table_ablationstudycity}. We achieved an accuracy of  up to 92.8\% on the Jigsaw task which 
supports our hypothesis that FCN is essentially a patch-wise classification network and can be used to solve the Jigsaw problem.

\subsection{Self-supervised Learning as Proxy Task}
\label{subsec_exp_semantic}
We also validate the effectiveness of our self-supervised learning by performing a transfer learning task, i.e., use the learned features as pre-trained weights for semantic segmentation.

Many works have proved that the self-supervised features at deeper layers are specific to the proxy tasks, and are hence not well suited for downstream tasks. To find the best transfer strategy between our self-supervised learning and semantic segmentation, we propose to freeze a subset of layers during the finetune process. In addition, the unfrozen layers were initialized either by random values or self-supervised features. 
Specifically, since most CNNs can be divided into five blocks according to the resolution of feature maps, we perform weights freezing using the ``block" as unit (see Table \ref{table_ablationstudycity}). Note that the final classification layer of segmentation is always initialized with random values since its dimension is different from the one used in self-supervised learning.  To validate how the performance of proxy task influences the performance of segmentation task, we further perform transfer learning using parameters saved at different training steps. For convenience, we call these self-supervised models by PARAM-AT-N, where ``N" is the training steps (iterations).

To train the segmentation network, we set the batch size to 8 and the training image size to $1024\times1024$. The total training steps was set to 30K. The learning rate was set to 0.05 and decayed by a factor of 10 at 5K, 15K, and 25K steps. Other training policies were similar to those used in the self-supervised learning.

 We used the commonly used mean Intersection over Union (mIoU) metric to evaluate the performance of semantic segmentation. Results of our ablation experiments are shown in Table \ref{table_ablationstudycity}. From the Table, important observations and conclusions can be drawn as follows:

\begin{enumerate}
\item The baseline method, the entire feature extraction network was learned from random values with full supervision, achieved 42.0\% mIoU on the Cityscapes validation set. When only block4 and block5 were learned from random values, our method, no matter the self-supervised models used,  achieved comparable or better performance than the baseline. In addition, segmentation models initialized from our self-supervised features (finetune columns) always performed better than those initialized from random values (random columns). These results show that our self-supervised learning can learn useful feature representations.

\vspace{-2mm}
\item Using PARAM-AT-30K for transfer learning,  we obtained a 1.2 percentage point improvement over the baseline when all feature blocks (block12345) were finetuned. The improvements achieved over the baseline increased first and then decreased with the increase in the number of frozen blocks. In particular, we arrived at the inflection point and obtained the largest improvement (2.8 percentage point) over the baseline when block1 and block2 were frozen. However, the segmentation performance decreased significantly to 27.2\% mIoU when block1 to block4 where all frozen (only block5 was not frozen). These results indicate that the low-level features learned by self-supervised learning are generic to downsteam tasks while high-level features are specific to the proxy task (Jigsaw in our case). Note that this and the following observations/conclusions are based on the results listed in the ``finetune" columns of the Table \ref{table_ablationstudycity}. 

\vspace{-2mm}
\item Compared to using PARAM-AT-30K, the inflection point of performance came earlier when using PARAM-AT-50K. This suggests that the low-level features will also become specific to proxy task with the increase of training steps of self-supervised learning.

\vspace{-2mm}
\item The results obtained with PARAM-AT-20K seemed to conflict with the third observation since the best performance also came earlier than using PARAM-AT-30K. Nevertheless, since the reduction in mIoU were both larger than one percentage point after the inflection points of PARAM-AT-30K and PARAM-AT-50K, we believe the inflection point of PARAM-AT-20K was the same with the one of PARAM-AT-30K.

\vspace{-2mm}
\item When block1 and block2 were frozen, PARAM-AT-20K outperformed PARAM-AT-50K but performed worse than PARAM-AT-30K. A possible explanation is that the block2 of PARAM-AT-20K was neither good enough for the semantic segmentation nor for the proxy task. Combined with the third point, it can be concluded that the training steps of self-supervised learning should not be too large or too small for transfer learning.
\end{enumerate}

Overall, the above experiments strongly suggest that our self-supervised learning can learn useful features, especially at the low-level, for semantic segmentation.

\subsection{Jigsaw Puzzle With More Patches}
\label{subsec_exp_25patches}
Following the work of ~\cite{noroozi2016unsupervised}, we adopted nine patches for our Jigsaw-based self-supervised learning in the above experiments. In addition, these nine patches were sampled from an image of spatial size $576\times576$, which means each patch was $192\times192$. In practice, the number of patches of a Jigsaw Puzzle can be increased to hundreds or even thousands and the patch size can also be varied. Thus, to test if our method can learn better representations by solving a Jigsaw Puzzle with more patches, we increase the patches to 25 in this experiment. We also experimented with different patch sizes to evaluate how this influences the performance of our self-supervised learning.

For the self-supervised learning, we first adopted the same training image size as the one used in 9 patches, leading to 25 patches of size $115\times115$. Then, we adopted the same patch size as the one used in 9 patches and the training image size thus became $960\times960$. Note that each convolution layer of MobileNetV2 is followed by a batch normalization layer that performs feature normalization for each channel over all the training pixels. In other words, the performance of batch normalization can be easily affected by the total number of training pixels. Thus, for a fair comparison, we adjusted the batch size to 13 when the training image size was $960\times960$ to ensure the total training pixels stay approximately the same to those used in other settings. All other training policies were kept the same with those depicted in Section \ref{subsec_selfastarget}.

When transferred to semantic segmentation, all the training policies are kept the same as depicted in Section \ref{subsec_exp_semantic} and the self-supervised models used for transferring were ``PARAM-AT-30K". The experimental results are reported in Table \ref{table_25patches}, where ``rd'' and ``ft'' represents random and fine tune, respectively.

\begin{table}[ht]
\scriptsize%
\begin{center}
\begin{tabular}{ccc|cc|cc|cc}
\hline
\multirow{2}{*}{Patches}& \multirow{2}{*}{sizes}&\multirow{2}{*}{ACC} &\multicolumn{2}{c}{block345}&\multicolumn{2}{c}{block45}&\multicolumn{2}{c}{block5}\\
   &   &   &rd & ft &rd & ft &rd & ft\\

\hline
$3\times3$ &192 & 85.1 &42.2  & 44.8  &42.2 & 43.2  & 26.8 &27.5\\
\hline
\multirow{2}{*}{$5\times5$}&115  &48.2  &43.0  &47.8  &43.5 &46.2   & 31.7 &32.8\\
           &192 &56.6  & -  & &- &   & - &33.1\\

\hline
\end{tabular}
\end{center}
\caption{Ablation study on patch numbers and sizes with the unfrozen block parameters initialized randomly `rd' or fine tuned `ft' form our self-supervised model.}
\label{table_25patches}
\end{table}

\begin{figure}[t]
\begin{center}
   \includegraphics[width=0.8\linewidth]{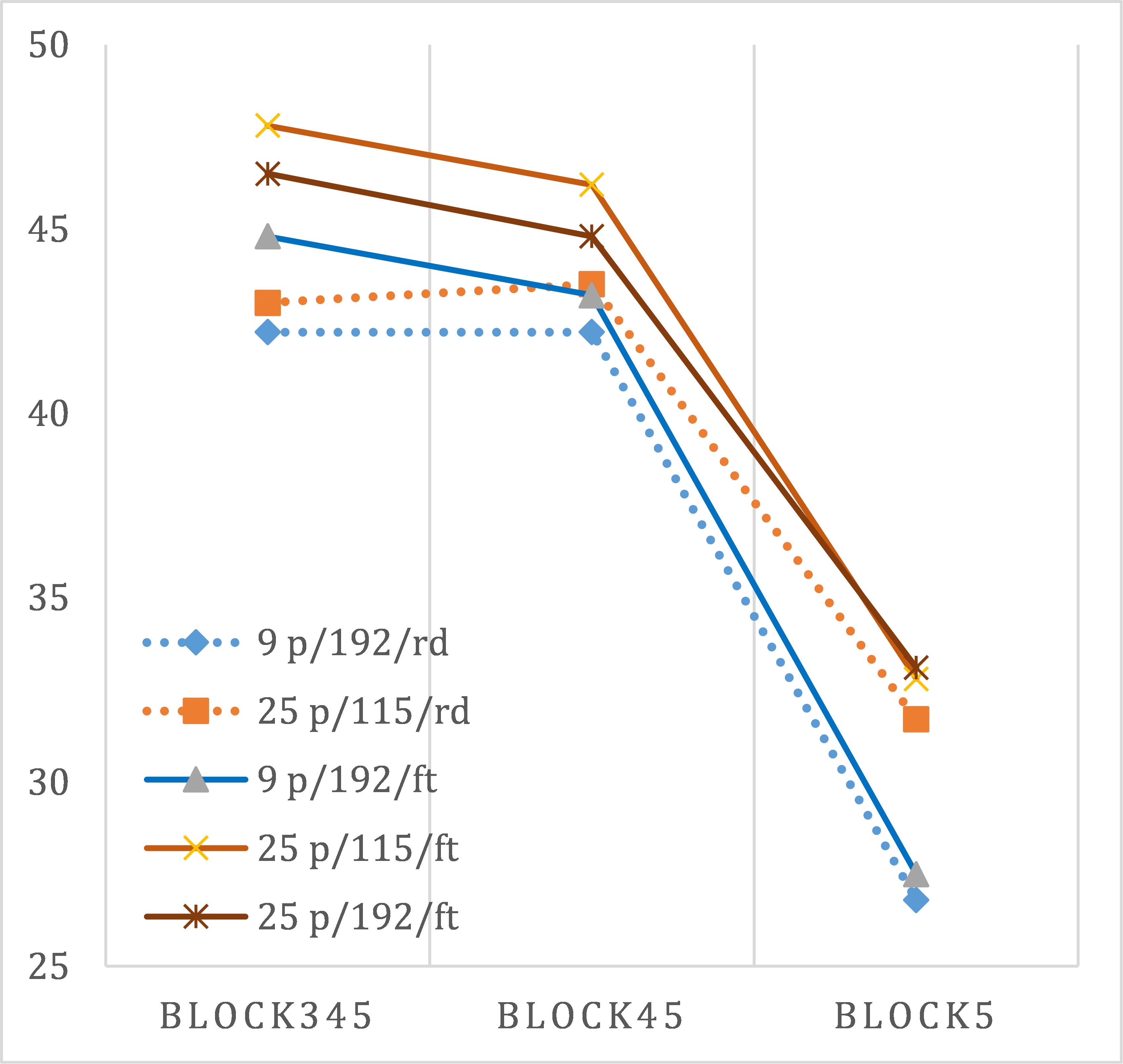}
\end{center}
   \caption{Transfer learning accuracy (vertical axis) using our self-supervised models trained with different patch numbers and sizes. The dotted and solid lines mean that the parameters of unfrozen blocks (shown on the horizontal axis) are initialized from random values and self-supervised features, respectively. }
\label{fig_25patches}
\end{figure}

\begin{table*}[]
\footnotesize

\begin{center}
\begin{tabular}{lllll}

\hline
\multirow{2}{*}{Method } & \multirow{2}{*}{Backbone}  & \multicolumn{2}{c}{Pre-training Setup} & \multirow{2}{*}{mIoU} \\
\cline{3-4}
 & &Dataset & epoches &\\
\hline
ImageNet-Labels~\cite{krizhevsky2012imagenet}       &AlexNet    &ImageNet(1.28M)   &            & 48.0      \\
\hline
Random Gaussian      &   AlexNet  & -   &  -         &    19.8~\cite{pathak2016context}   \\
Autoencoder       &  AlexNet  & -    &  -         &    25.2~\cite{pathak2016context}   \\

Kr{\"a}henb{\"u}hl et al. ~\cite{krahenbuhl2015data}      &  AlexNet     &ImageNet      &-        &32.6~\cite{donahue2016adversarial}   \\
\hline
Inpainting~\cite{pathak2016context}             &  ALexNet+BN    & ImageNet           &10    &   30.0   \\

Counting~\cite{noroozi2017representation}     &AlexNet          &ImageNet    &-      & 36.6      \\
Colorization ~\cite{zhang2016colorful}*        &AlexNet+BN        & ImageNet  & -          & 35.6           \\
Split-Brain~\cite{zhang2017split}*             & AlexNet+BN       &ImageNet   &-          & 36.0          \\
Jigsaw~\cite{noroozi2016unsupervised}           & AlexNet        &ImageNet     &  70      &  37.6    \\
Spot-Artifacts~\cite{jenni2018self}*          & AlexNet      & ImageNet      & -        &  38.1    \\
Colorization ~\cite{larsson2017colorization}  &AlexNet+BN+Hypercolumn  & ImageNet+Places(2.4M)~\cite{zhou2014learning} &  3-10      & 38.4     \\
RotNet~\cite{gidaris2018unsupervised}*         & AlexNet+BN       &ImageNet   & 30      & 39.1     \\
Mundhenk et al.~\cite{nathan2018improvements}  & AlexNet     &ImageNet  &  150       &  41.4~\cite{feng2019self}     \\
\hline
Ours      & AlexNet     &  PASCAL VOC Test(16K)       & 200     & 38.0    \\
\hline
\end{tabular}

\end{center}
\caption{Comparing our method to SOTA methods on PASCAL VOC2012 validation set. * indicates the use of data-dependent re-scaling method proposed by  Kr{\"a}henb{\"u}hl et al. ~\cite{krahenbuhl2015data}. The mIoU values with citations means they are excerpted from the corresponding cited paper. AlexNet+BN represents that each linear layer of AlexNet is followed by a batch normalization layer.}
\label{exp_comparison-to-sota}
\end{table*}

\noindent {\bf Analysis:} Not surprisingly, it can be seen from the Table \ref{table_25patches} that the accuracy of Jigsaw task reduced significantly when the patch number increased from 9 to 25. Nevertheless, the performance of segmentation task improved significantly in this case. For example,  the segmentation accuracy improved by 4.9 percentage points when block 1,2,3,4 were frozen and the block5 was initialized from random values. This suggests that our Jigsaw-based self-supervised learning can learn better high-level semantic features with 25 patches. One reasonable explanation is that it becomes more difficult to exploit shortcut solutions to solve the Jigsaw puzzle when more patches are used. The shortcut solutions may learn information highly specific to the Jigsaw puzzle task but not the target task, i.e., the semantic segmentation. As mentioned by Noroozi et al., \cite{noroozi2016unsupervised}, the shortcuts to solve the Jigsaw puzzle task mainly include low-level statistics, such as edge continuity, the pixel intensity/color distribution, and chromatic aberration. Among these, the Chromatic aberration, which is
relatively difficult to understand, is a relative spatial shift between color channels that increases from the images center to the borders. Noroozi et al. propose to avoid these three shortcuts by using patch-independent normalization, overlapped patches, and color jittering, respectively.  Instead of introducing these explicit strategies, we believe our method can avoid these shortcuts implicitly. Note that the FCN can be viewed approximately, but not exactly, as a patch-wise classification framework. In other words, the patches of Jigsaw Puzzle in our method are implicitly overlapped although the overlapped area is limited. Nevertheless, the overlap area would increase with the increase of patch numbers given the same image size.


%


\subsection{Comparison to Other Methods}
We finally evaluated our self-supervised learning on PASCAL VOC2012 to compare it with other methods. The original PASCAL VOC2012 training set contains only 1,464 dense annotated images for semantic segmentation.  Hariharan et al.~\cite{hariharan2011semantic} then expanded it to 10,582 images. Existing methods usually performed the self-supervised learning on ImageNet~\cite{deng2009imagenet} that contains about 1.3M images of 1000 natural classes, which could cost more than one week on a single Titan X GPU. In contrast, we performed the self-supervised learning on the 16,135 images included in the ``test\_JPEGImages" of PASCAL VOC2012. Similar to most methods, we also adopt the AlexNet, which consists of five convolution layers and two fully connected layers, for feature learning in this experiment.


\noindent {\bf Implementation Details:} During self-supervised learning, we set the batch size to 64 and the training image size to $450\times450$. We train the model for 50K  steps (about 200 epochs). The initial learning rate was set to 0.01 and then decayed by a factor of 10 at 20K and 40K steps. In the context of semantic segmentation, we trained the FCN32 model for 50K steps. The training batch was set to $36\times500\times500$. The initial learning rate was set to 0.01 and then decayed by a factor of 10 at 10K, 25K and 40K steps. Similar to ~\cite{shelhamer2017fully}, we also changed the padding number of the first convolution layer to 100 when transferring the learned features to semantic segmentation.  The experimental results are reported in Table \ref{exp_comparison-to-sota}.



\noindent {\bf Analysis:} When all the parameters were initialized from Random Gaussian values, which acts as the baseline in this experiment, the AlexNet-FCN32 achieved 19.8\% mIoU on the PASCAL VOC2012 validation set. In contrast, when all the parameters were pre-trained with ImageNet classification labels, the AlexNet-FCN32 achieved the ceiling value 48\% mIoU.  Our method achieved 38.0\% mIoU on the PASCAL VOC2012 validation set, which outperformed significantly the baseline and the auto-encoder method. This experiment and the above experiments on Cityscapes dataset strongly support our argument that our self-supervised learning method can be applied to different datasets and models. Compared with state-of-the-art methods, our method, without using batch normalization and data-dependent rescaling, outperformed Inpainting~\cite{pathak2016context}, Colorization ~\cite{zhang2016colorful}, Split-Brain~\cite{zhang2017split}, and Jigsaw method~\cite{noroozi2016unsupervised} by a large margin and achieved competitive performance with other methods using significant fewer pre-training images. In particular, our method trained on a dataset that contains only 16K images for 200 epochs, resulting in total 1.9M training images. In contrast,  the RotNet and Mundhenk's method use 68.4M and 192M training images, respectively.
%




\section{Conclusion}
\label{sec_conclusion}
In this paper, we presented a novel self-supervised learning framework for semantic segmentation. We showed that the classical FCN can be approximately viewed as a patch-wise classification framework and applied to solve a Jigsaw Puzzle problem for representation learning. We achieved 5.8\%  mIoU improvement over the baseline model that was initialized from random values on Cityscapes validation set. Moreover, we achieved competitive performance with state-of-the-arts on the PASCAL VOC2012 dataset with significant fewer pre-training image sources.



{\small
\bibliographystyle{ieee_fullname}
\bibliography{mybibfile_nodoi}
}

\end{document}